\documentclass[review]{elsarticle}
\graphicspath{ {./figures/} }
\usepackage{hyperref}
\usepackage{float}
\usepackage{verbatim} 
\usepackage{apalike}
\restylefloat{figure}
\usepackage{rotating} 

\graphicspath{ {./figures/} }
\usepackage[table]{xcolor}
\usepackage{verbatim} 
\usepackage{apalike}
\usepackage{caption}
\usepackage{subcaption}
\usepackage[margin=1in]{geometry} 
\usepackage[]{graphicx}
\usepackage[font=small,skip=0pt]{caption}
\usepackage{enumitem}
\usepackage{booktabs}
\usepackage{longtable}
\usepackage{tabularx}
\usepackage{cancel}
\usepackage{multirow}
\usepackage{supertabular}
\usepackage{algorithmic}
\usepackage{algorithm}
\usepackage{amsmath}
\usepackage{rotating}
\usepackage[normalem]{ulem}
\usepackage[utf8]{inputenc}
\usepackage[T1]{fontenc}
\usepackage{lineno}
\usepackage{array}
\usepackage{makecell}
\usepackage{pdflscape}
\usepackage{afterpage}
\usepackage{capt-of}
\usepackage{lipsum}
\usepackage{changepage}
\usepackage{xcolor}
\usepackage{rotating} 
\usepackage{graphicx}
\newcommand{\blue}[1]{\textcolor{blue}{#1}}
\usepackage{amssymb}
\usepackage{natbib}
\bibpunct{(}{)}{,}{a}{}{;}
\renewcommand{\citep}[1]{(\citealp{#1})}

\biboptions{authoryear}

\begin{document}

\begin{frontmatter}

\title{ Recognition of Harmful Phytoplankton from Microscopic Images using Deep Learning }

\author[label2]{Aymane Khaldi}
\ead{ay.khaldi@aui.ma}
\author[label1]{Rohaifa Khaldi \corref{cor1}}
\ead{rohaifa@ugr.es}

\cortext[cor1]{Corresponding author.}
\address[label1]{Dept. of Computer Science and Artificial Intelligence, DaSCI, University of Granada, 18071 Granada, Spain}
\address[label2]{School of Science and Engineering, Al Akhawayn University, Ifrane, Morocco}

\begin{abstract}
Monitoring plankton distribution, particularly harmful phytoplankton, is vital for preserving aquatic ecosystems, regulating the global climate, and ensuring environmental protection. Traditional methods for monitoring are often time-consuming, expensive, error-prone, and unsuitable for large-scale applications, highlighting the need for accurate and efficient automated systems. In this study, we evaluate several state-of-the-art CNN models, including ResNet, ResNeXt, DenseNet, and EfficientNet, using three transfer learning approaches: linear probing, fine-tuning, and a combined approach, to classify eleven harmful phytoplankton genera from microscopic images. The best performance was achieved by ResNet-50 using the fine-tuning approach, with an accuracy of 96.97\%. The results also revealed that the models struggled to differentiate between four harmful phytoplankton types with similar morphological features.

\end{abstract}


\end{frontmatter}

\section{Introduction}\label{S1}

Phytoplankton serves as a vital foundation for life in aquatic ecosystems and plays a key role in energy transfer within the aquatic food web \citep{rachman2022application}. Plankton is responsible for producing approximately 90\% of the Earth's oxygen, making its low levels a threat to ocean ecosystems. Conversely, excessive plankton growth can have destructive consequences due to toxin production \citep{zhao2010binary}. 

Due to its rapid response and high sensitivity to environmental changes, many phytoplankton species are used as indicators of environmental health such as global warming \citep{lumini2019deep}.
The growth rate, the cell shape, and other metabolites can indicate various anomalies in the water, such as eutrophication. Additionally, certain phytoplankton species tend to proliferate rapidly under abnormal water conditions, leading to harmful algal blooms (HABs), which can have severe impacts on the ecosystem, particularly if the blooming species are toxin producers \citep{suthers2019importance}.
Furthermore, harmful phytoplankton species such as Aphanizomenon, Microcystis, Anabaena, and Oscillatoria are common cyanobacteria \citep{paerl2001harmful}, some of which produce hepatotoxins and neurotoxins that pose direct threats to water supplies and the safety of both animals and humans.


Studying and monitoring plankton distribution and particularly harmful phytoplankton is crucial for the maintenance of the aquatic ecological environment, global climate regulation, and environmental protection.
Traditional algae monitoring methods rely on highly experienced professionals, making them time-consuming, costly, error-prone, and often impractical for large-scale use \citep{yang2023automatic}. As a result, researchers are increasingly turning to automated approaches for plankton classification using computer vision techniques.

Several studies have leveraged deep learning, particularly Convolutional Neural Networks (CNNs), to automate plankton image acquisition, identification, and enumeration \citep{rachman2022application}. For instance, \cite{yang2023automatic} employed models such as AlexNet, VGG16, GoogLeNet, ResNet50, and MobileNetV2 to identify harmful phytoplankton, while \cite{rachman2022application} utilized VGG16 to classify various phytoplankton species. Additionally, \cite{figueroa2024phytoplankton} applied Faster R-CNN and RetinaNet for detecting freshwater phytoplankton, and \cite{kyathanahally2021deep} used multiple CNN models with ensemble learning to recognize zooplankton in lakes.

The primary objective of this study is to develop an automated system for recognizing toxic phytoplankton using several state-of-the-art CNN models, including ResNet, ResNeXt, DenseNet, and EfficientNet. We also evaluate three different transfer learning approaches: linear probing, fine-tuning, and a combined approach. The rest of the paper is organized as follows: Section \ref{S2} details the experimental setup, Section \ref{S3} presents the experimental results, and Section \ref{S4} provides a summary of the study.

\section{Experiment}\label{S2}
In this section, we present the phytoplankton dataset, as well as the different learning approaches used to develop the automatic system for phytoplankton recognition, finally, we present the metrics used for model evaluation.

\subsection{Data Acquisition}
\begin{figure}[H]
    \centering
    \includegraphics[width=13cm]{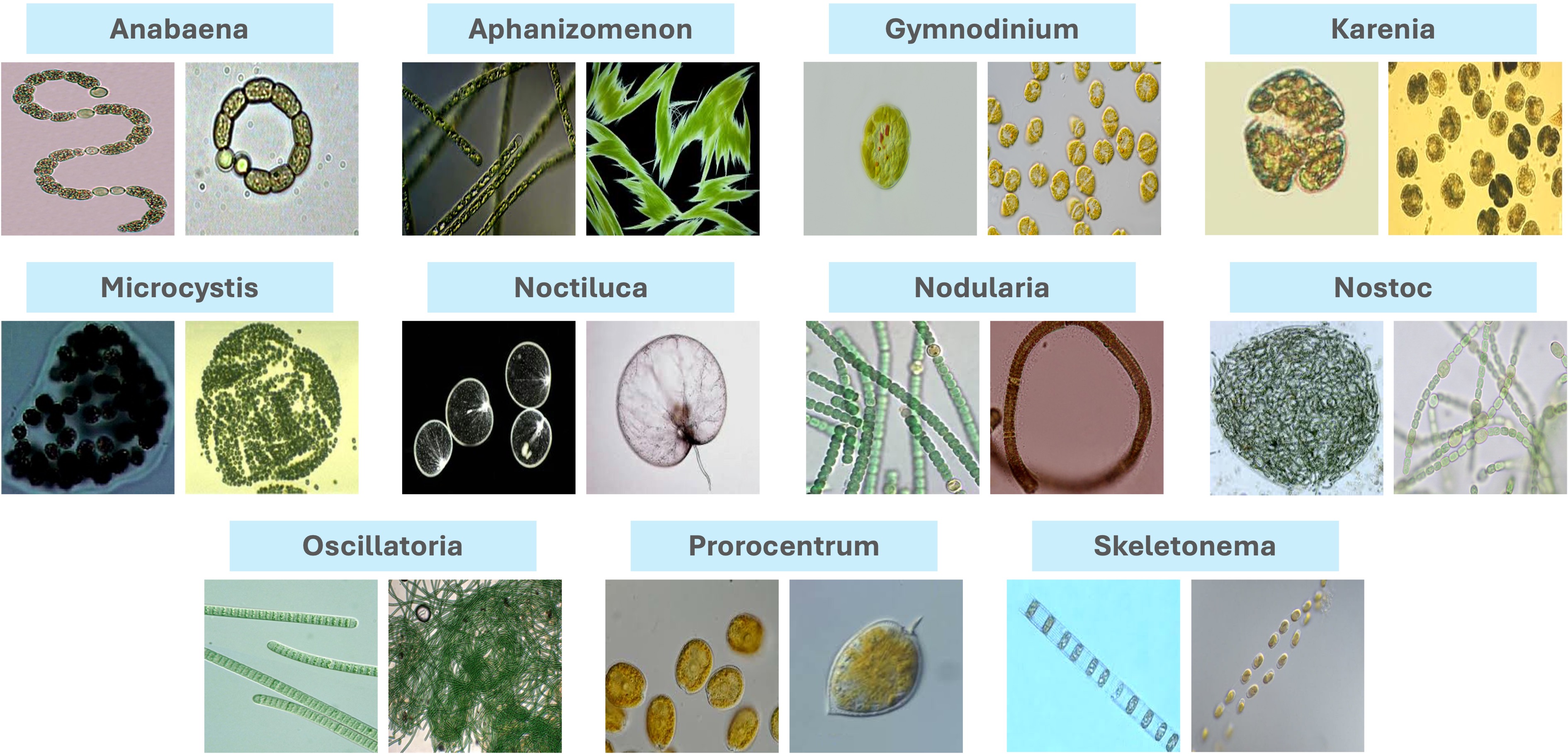}
    \caption {Visualization of two samples per class from the toxic phytoplankton dataset.}
    \label{data}
\end{figure}

In this study, we utilized a publicly available dataset of phytoplankton \citep{yang2023automatic}. This dataset includes eleven common harmful phytoplankton genera, encompassing a total of 1650 images, both original and augmented (see Fig. \ref{data}). Each class is represented by 150 RGB images, which were resized to 224x224 pixels. 
The dataset used in this study was downloaded from the web page \href{https://www.kaggle.com/datasets/mengyuy/the-algae-cell-images}{\blue{(the-algae-cell-images)}}.
To perform the experiment, the dataset was partitioned into three subsets: (1) training data for model parameter optimization, (2) validation data to select the best model configuration and avoid overfitting, and (3) test data to evaluate the model generalization performance.

\subsection{Experimental Design}
\begin{figure}[H]
     \centering
     \begin{subfigure}[b]{0.4\textwidth}
         \centering
         \includegraphics[width=7cm]{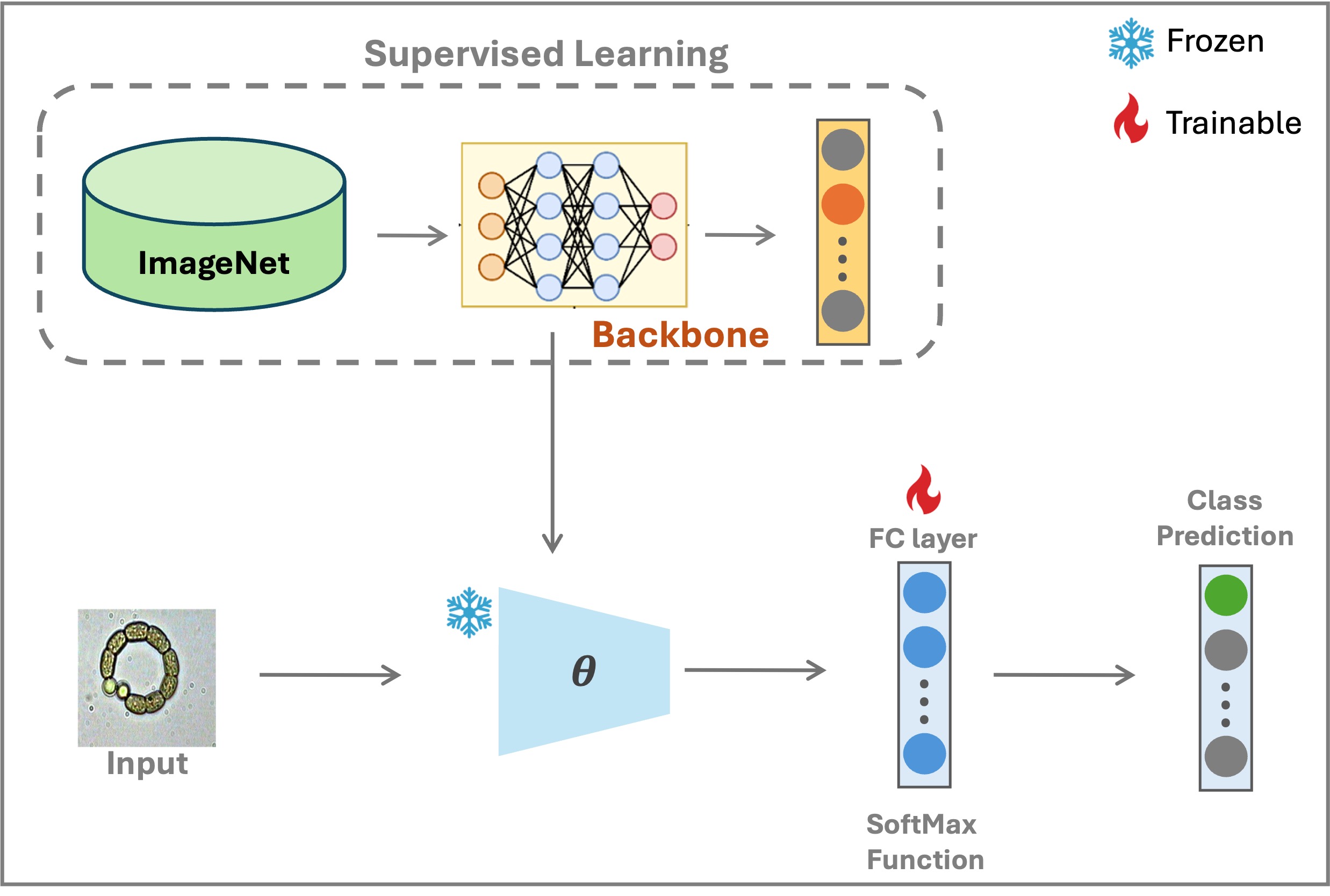}
         \caption{}
         \label{d1}
     \end{subfigure}
     \hfill
     \begin{subfigure}[b]{0.5\textwidth}
         \centering
         \includegraphics[width=7cm]{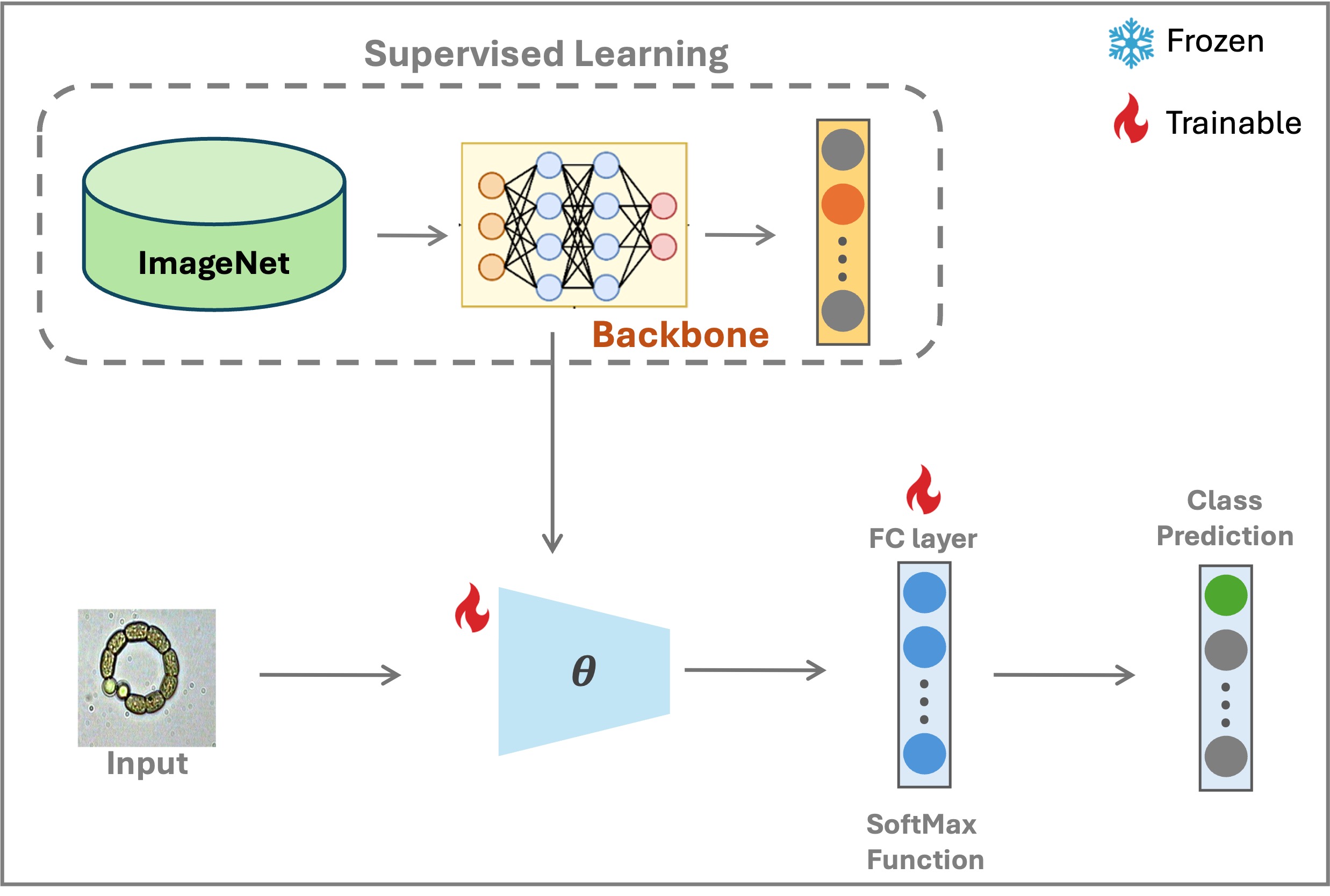}
         \caption{}
         \label{d2}
     \end{subfigure}
     \hfill
     \begin{subfigure}[b]{0.5\textwidth}
         \centering
         \includegraphics[width=7cm]{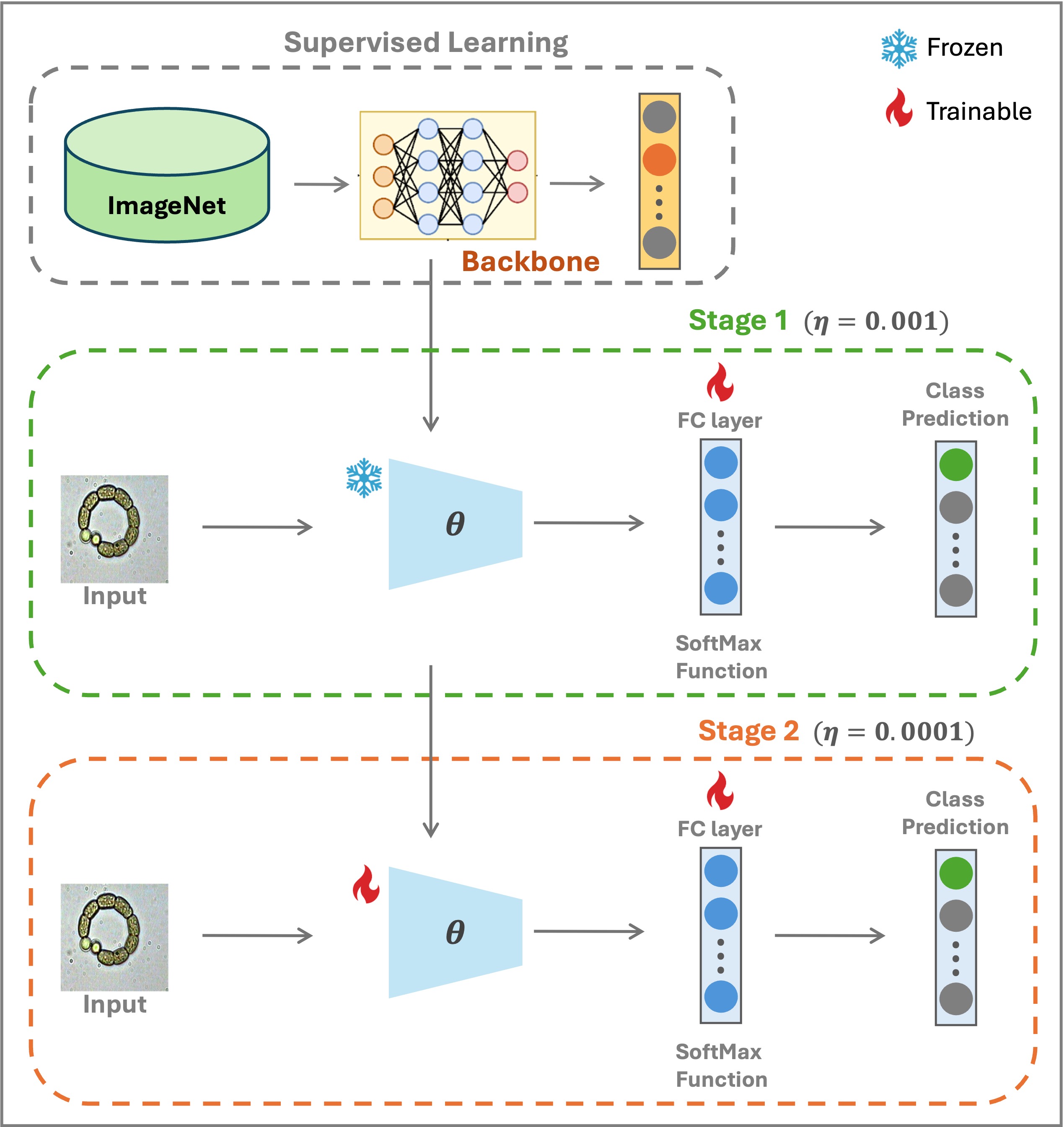}
         \caption{}
         \label{d3}
     \end{subfigure}
        \caption{Overview of the three learning approaches employed for developing the automatic phytoplankton recognition system: (a) Learning approach using linear probing, (b) Learning approach utilizing fine-tuning, and (c) Learning approach that integrates both linear probing and fine-tuning.}
        \label{exp}
\end{figure}

In this study, we employed transfer learning to develop an automatic phytoplankton recognition system (see Fig. \ref{exp}). We explored three distinct learning approaches: (1) Linear Probing, where the model backbone was frozen, and only the classifier head (i.e., the final classification layer) was trained for 100 epochs (Fig. \ref{d1}). (2) Fine-Tuning, where both the backbone and the classifier head were trained for 100 epochs (Fig. \ref{d2}). (3) Combined Approach, which integrates both linear probing and fine-tuning into a two-stage process. Initially, the model was trained with linear probing for 50 epochs at an initial learning rate of 0.001, followed by fine-tuning for an additional 50 epochs at a reduced learning rate of 0.0001 (Fig. \ref{d3}).

For this experiment, we evaluated several CNN backbones, varying batch sizes and initial learning rates. All models were pretrained on ImageNet using the Adam optimizer, with their classifier heads adapted to address our multiclass classification problem involving eleven classes.

\subsection{Model Evaluation}
To evaluate the overall performance of the model, we employed three metrics: precision (\ref{eq2}) to measure the accuracy of the model in recognizing different phytoplankton genera, recall (\ref{eq3}) to assess the model’s effectiveness in retrieving instances of each genus, and accuracy (\ref{eq1}), which indicates the proportion of correctly classified images relative to the total number of images.

\begin{equation}
    \text{Precision} = \frac{\text{TP}}{\text{TP+FP}}
    \label{eq2}
\end{equation}

\begin{equation}
    \text{Recall} = \frac{\text{TP}}{\text{TP+FN}}
    \label{eq3}
\end{equation}

\begin{equation}
    \text{Accuracy} = \frac{\text{TP+TN}}{\text{TP+FP+TN+FN}}
    \label{eq1}
\end{equation}

\section{Results and Discussion}\label{S3}

\begin{figure}[H]
  \begin{center}
    \includegraphics[width=9cm]{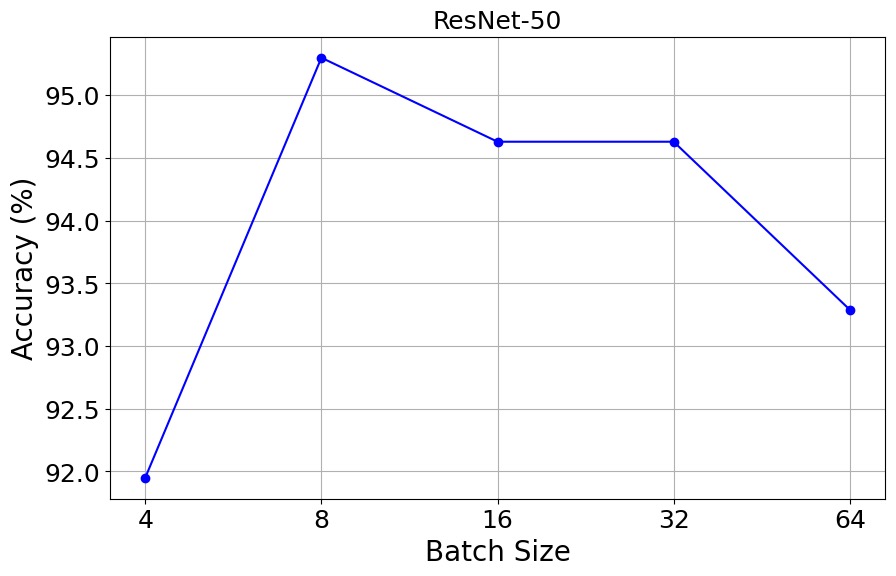} 
  \end{center}
  \caption{ResNet-50 accuracy on validation set across various batch sizes using a learning strategy that combines linear probing and fine-tuning.}
  \label{bs}
\end{figure}

Fig. \ref{bs} shows the validation results of training the ResNet-50 model using a transfer learning approach with a combined strategy, evaluated across various batch sizes (4, 8, 16, 32, and 64). The optimal performance, with an accuracy of 95.3\%, was achieved using a batch size of 8 images.

Table \ref{tab1} displays the validation results of training the ResNet-50 model with a batch size of 8 images across three different learning approaches, each with varying initial learning rates. The highest performance, with an accuracy of 95.97\%, was obtained using the fine-tuning approach with an initial learning rate of 0.0001.
Table \ref{tab2} presents the test results for various CNN backbones trained using the fine-tuning approach with a batch size of 8 images and an initial learning rate of 0.0001. The best performance was achieved by ResNet-50, which attained an accuracy of 96.97\%, a precision of 96.99\%, a recall of 96.97\%, and a training time of 18.51 minutes.

\begin{table}[H]
\centering
\caption{Evaluation of ResNet-50 on validation set across different learning designs trained with a batch size of 8 images.}
\label{tab1}
\begin{tabular}{llr} \hline
\multicolumn{1}{c}{\textbf{Transfer learning approach}} & \multicolumn{1}{c}{\textbf{\thead{Learning rate}}} & \multicolumn{1}{c}{\textbf{Accuracy (\%)}} \\ \hline
Linear probing then fine-tuning & 0.001 - 0.0001 & 95.3 \\
Linear probing    & 0.001          & 93.29                    \\
Linear probing    & 0.0001         & 92.62                    \\
Fine-tuning    & 0.001          & 89.26                    \\
\textbf{Fine-tuning}    & \textbf{0.0001}         & \textbf{95.97}              \\   
\hline
\end{tabular}
\end{table}

\begin{table}[H]
\centering
\caption{Evaluation of various CNN backbones on test data for phytoplankton recognition, trained with fine-tuning and a batch size of 8 images.}
\label{tab2}
\begin{tabular}{lrrrr}
\hline
\textbf{Backbone} & \multicolumn{1}{c}{\textbf{\thead{Accuracy\\(\%)}}} & \multicolumn{1}{c}{\textbf{\thead{Precision\\(\%)}}} & \multicolumn{1}{c}{\textbf{\thead{Recall\\(\%)}}} & \multicolumn{1}{c}{\textbf{\thead{Training time \\ (min)}}} \\ \hline
ResNet-18       & 93.33 & 93.56 & 93.33 & 12.75 \\
\textbf{ResNet-50}       & \textbf{96.97} & \textbf{96.99} & \textbf{96.97} & \textbf{18.51} \\
ResNet-152      & 95.15 & 95.58 & 95.15 & 35.5  \\
ResNeXt-50      & 95.76 & 95.82 & 95.76 & 21.97 \\
DenseNet-121    & 91.52 & 92.04 & 91.52 & 27.47 \\
EfficientNet-B0 & 96.36 & 96.47 & 96.36 & 20.87 \\
\hline
\end{tabular}
\end{table}

\begin{figure}[H]
  \begin{center}
    \includegraphics[width=11cm]{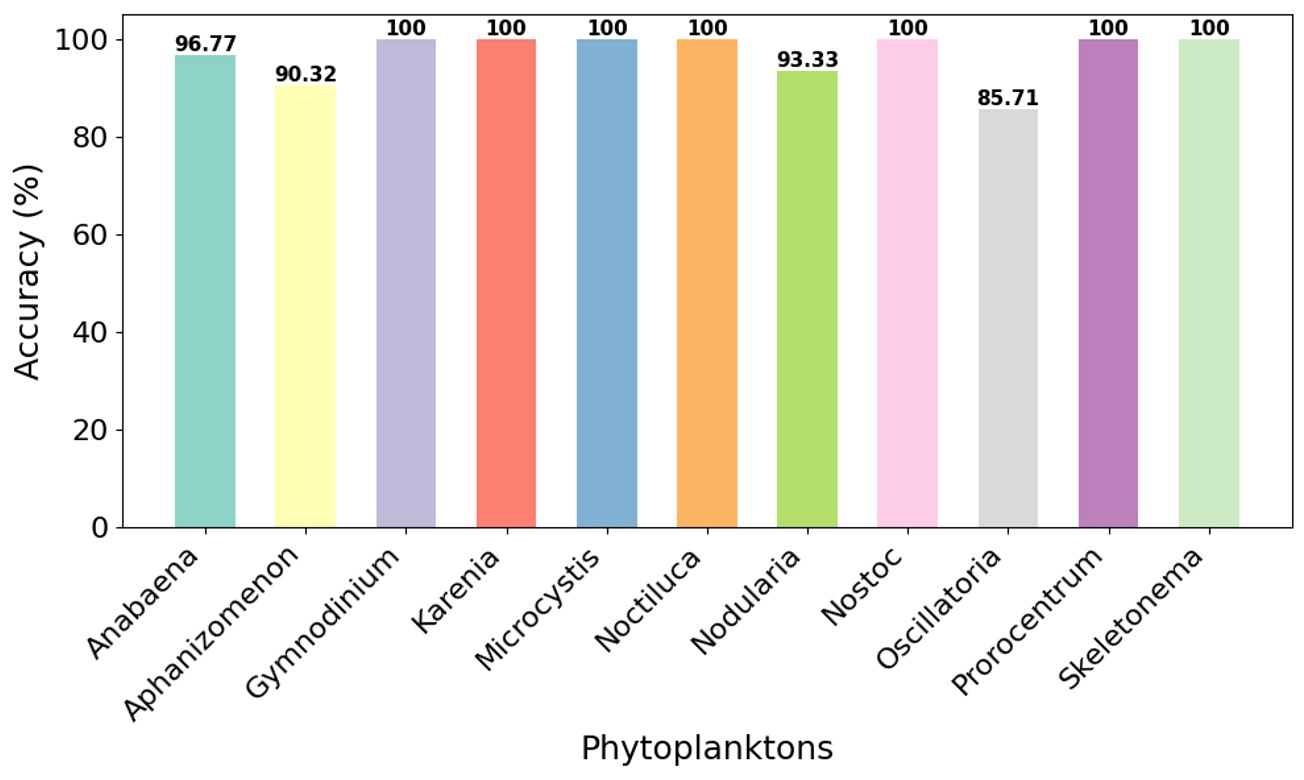} 
  \end{center}
  \caption{Evaluation of ResNet-50 on test data for different types of toxic phytoplankton, trained using fine-tuning.}
  \label{fig4}
\end{figure}

\begin{figure}[H]
  \begin{center}
    \includegraphics[width=11cm]{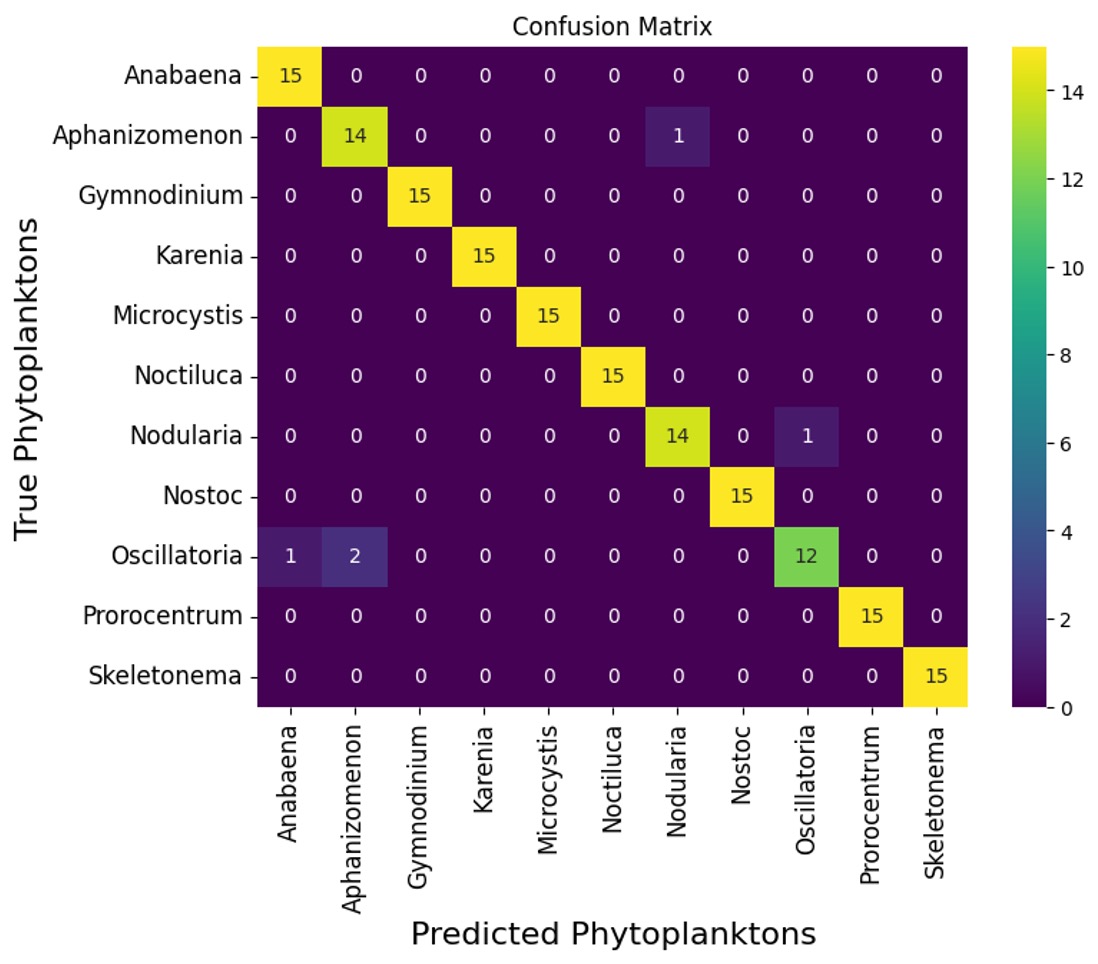} 
  \end{center}
  \caption{Confusion matrix displaying true and predicted phytoplankton class distributions on test data, generated by ResNet-50 trained with fine-tuning.}
  \label{fig5}
\end{figure}

Fig. \ref{fig4} illustrates the test performance of the top-performing model, ResNet-50, across various phytoplankton genera. The model achieved an accuracy of 100\% for all genera except \textit{Oscillatoria}, which had an accuracy of 85.71\%, \textit{Aphanizomenon} with 90.32\%, \textit{Nodularia} at 93.33\%, and \textit{Anabaena} with 96.77\%. 
Fig. \ref{fig5} presents the confusion matrix for the top-performing model on the test data, highlighting several points of confusion: \textit{Aphanizomenon} is often misclassified as \textit{Nodularia}, \textit{Nodularia} is frequently mistaken for \textit{Oscillatoria}, and \textit{Oscillatoria} is commonly confused with both \textit{Aphanizomenon} and \textit{Anabaena}.
This is primarily due to the similar morphological patterns of these four genera of phytoplankton.

\section{Conclusion}\label{S4}
In this study, we evaluated six CNN architectures (ResNet-18, ResNet-50, ResNet-152, ResNeXt-50, DenseNet-121, and EfficientNet-B0) using three transfer learning approaches—linear probing, fine-tuning, and a combined approach—to classify eleven harmful phytoplankton genera from a publicly available benchmark dataset of microscopic images. The best performance was achieved by ResNet-50 with the fine-tuning approach, yielding an accuracy of 96.97\%. However, the models had difficulty distinguishing between four phytoplankton types with similar morphological characteristics. The proposed system offers a valuable tool for experts by automating species identification and reducing manual effort. Nonetheless, the development of a more comprehensive dataset encompassing a wider range of harmful phytoplankton is recommended to support the creation of more general models for water quality monitoring.

\section*{Declaration of Competing Interest}
The authors declare that they have no known competing financial interests or personal relationships that could have appeared to influence the work reported in this paper.


\bibliography{sample}

\begin{thebibliography}{8}
\expandafter\ifx\csname natexlab\endcsname\relax\def\natexlab#1{#1}\fi
\providecommand{\url}[1]{\texttt{#1}}
\providecommand{\href}[2]{#2}
\providecommand{\path}[1]{#1}
\providecommand{\DOIprefix}{doi:}
\providecommand{\ArXivprefix}{arXiv:}
\providecommand{\URLprefix}{URL: }
\providecommand{\Pubmedprefix}{pmid:}
\providecommand{\doi}[1]{\href{http://dx.doi.org/#1}{\path{#1}}}
\providecommand{\Pubmed}[1]{\href{pmid:#1}{\path{#1}}}
\providecommand{\bibinfo}[2]{#2}
\ifx\xfnm\relax \def\xfnm[#1]{\unskip,\space#1}\fi
\bibitem[{Figueroa et~al.(2024)Figueroa, Rivas-Villar, Rouco \& Novo}]{figueroa2024phytoplankton}
\bibinfo{author}{Figueroa, J.}, \bibinfo{author}{Rivas-Villar, D.}, \bibinfo{author}{Rouco, J.}, \& \bibinfo{author}{Novo, J.} (\bibinfo{year}{2024}).
\newblock \bibinfo{title}{Phytoplankton detection and recognition in freshwater digital microscopy images using deep learning object detectors}.
\newblock {\it \bibinfo{journal}{Heliyon}\/},  {\it \bibinfo{volume}{10}\/}.
\bibitem[{Kyathanahally et~al.(2021)Kyathanahally, Hardeman, Merz, Bulas, Reyes, Isles, Pomati \& Baity-Jesi}]{kyathanahally2021deep}
\bibinfo{author}{Kyathanahally, S.~P.}, \bibinfo{author}{Hardeman, T.}, \bibinfo{author}{Merz, E.}, \bibinfo{author}{Bulas, T.}, \bibinfo{author}{Reyes, M.}, \bibinfo{author}{Isles, P.}, \bibinfo{author}{Pomati, F.}, \& \bibinfo{author}{Baity-Jesi, M.} (\bibinfo{year}{2021}).
\newblock \bibinfo{title}{Deep learning classification of lake zooplankton}.
\newblock {\it \bibinfo{journal}{Frontiers in microbiology}\/},  {\it \bibinfo{volume}{12}\/}, \bibinfo{pages}{746297}.
\bibitem[{Lumini \& Nanni(2019)}]{lumini2019deep}
\bibinfo{author}{Lumini, A.}, \& \bibinfo{author}{Nanni, L.} (\bibinfo{year}{2019}).
\newblock \bibinfo{title}{Deep learning and transfer learning features for plankton classification}.
\newblock {\it \bibinfo{journal}{Ecological informatics}\/},  {\it \bibinfo{volume}{51}\/}, \bibinfo{pages}{33--43}.
\bibitem[{Paerl et~al.(2001)Paerl, Fulton~III, Moisander \& Dyble}]{paerl2001harmful}
\bibinfo{author}{Paerl, H.~W.}, \bibinfo{author}{Fulton~III, R.~S.}, \bibinfo{author}{Moisander, P.~H.}, \& \bibinfo{author}{Dyble, J.} (\bibinfo{year}{2001}).
\newblock \bibinfo{title}{Harmful freshwater algal blooms, with an emphasis on cyanobacteria.}
\newblock {\it \bibinfo{journal}{TheScientificWorldJournal}\/},  {\it \bibinfo{volume}{1}\/}, \bibinfo{pages}{76--113}.
\bibitem[{Rachman et~al.(2022)Rachman, Suwarno \& Nurdjaman}]{rachman2022application}
\bibinfo{author}{Rachman, A.}, \bibinfo{author}{Suwarno, A.~S.}, \& \bibinfo{author}{Nurdjaman, S.} (\bibinfo{year}{2022}).
\newblock \bibinfo{title}{Application of deep (machine) learning for phytoplankton identification using microscopy images}.
\newblock In {\it \bibinfo{booktitle}{7th International Conference on Biological Science (ICBS 2021)}\/} (pp. \bibinfo{pages}{213--224}).
\newblock \bibinfo{organization}{Atlantis Press}.
\bibitem[{Suthers et~al.(2019)Suthers, Richardson \& Rissik}]{suthers2019importance}
\bibinfo{author}{Suthers, I.~M.}, \bibinfo{author}{Richardson, A.~J.}, \& \bibinfo{author}{Rissik, D.} (\bibinfo{year}{2019}).
\newblock \bibinfo{title}{The importance of plankton}.
\newblock {\it \bibinfo{journal}{Plankton: a guide to their ecology and monitoring for water quality}\/},  (pp. \bibinfo{pages}{1--13}).
\bibitem[{Yang et~al.(2023)Yang, Wang, Gao, Zhao, Li, Yang, Li, Li, Cui, Zhang et~al.}]{yang2023automatic}
\bibinfo{author}{Yang, M.}, \bibinfo{author}{Wang, W.}, \bibinfo{author}{Gao, Q.}, \bibinfo{author}{Zhao, C.}, \bibinfo{author}{Li, C.}, \bibinfo{author}{Yang, X.}, \bibinfo{author}{Li, J.}, \bibinfo{author}{Li, X.}, \bibinfo{author}{Cui, J.}, \bibinfo{author}{Zhang, L.} et~al. (\bibinfo{year}{2023}).
\newblock \bibinfo{title}{Automatic identification of harmful algae based on multiple convolutional neural networks and transfer learning}.
\newblock {\it \bibinfo{journal}{Environmental Science and Pollution Research}\/},  {\it \bibinfo{volume}{30}\/}, \bibinfo{pages}{15311--15324}.
\bibitem[{Zhao et~al.(2010)Zhao, Lin \& Seah}]{zhao2010binary}
\bibinfo{author}{Zhao, F.}, \bibinfo{author}{Lin, F.}, \& \bibinfo{author}{Seah, H.~S.} (\bibinfo{year}{2010}).
\newblock \bibinfo{title}{Binary sipper plankton image classification using random subspace}.
\newblock {\it \bibinfo{journal}{Neurocomputing}\/},  {\it \bibinfo{volume}{73}\/}, \bibinfo{pages}{1853--1860}.

\end{thebibliography}

\end{document}